\documentclass[lettersize,journal]{IEEEtran}
\usepackage{amsmath,amsfonts}
\usepackage{algorithmic}
\usepackage{algorithm}
\usepackage{array}
\usepackage[caption=false,font=normalsize,labelfont=sf,textfont=sf]{subfig}
\usepackage{array}
\usepackage{textcomp}
\usepackage{stfloats}
\usepackage{url}
\usepackage{verbatim}
\usepackage{graphicx}
\usepackage{cite}
\usepackage{float}
\usepackage{amssymb}
\usepackage{mathtools}
\usepackage{booktabs}
\usepackage{multirow}
\usepackage{caption}

\begin{document}
\bstctlcite{IEEEexample:BSTcontrol}

\title{RS-HyRe-R1: A Hybrid Reward Mechanism to Overcome Perceptual Inertia for Remote Sensing Images Understanding}

\author{{Gaozhi~Zhou, Hu~He, Peng~Shen, Jipeng~Zhang, Liujue~Zhang, Linrui~Xu, Zeyuan~Wang, Ziyu~Li, Xuezhi~Cui, Wang~Guo} ~\IEEEmembership{~Member,~IEEE,}
	{Haifeng~Li}~\IEEEmembership{Senior~Member,~IEEE}%

	\thanks{This work was supported in part by the National Natural Science Foundation of China under Grant 42271481, in part by Using Computing Resources at the High-Performance Computing Platform of Central South University, and in part by the Science and Technology Innovation Program of Hunan Province 2025RC3011.}
	\thanks{Gaozhi Zhou and Hu He are with the School of Mechanical and Electrical Engineering, Central South University, Changsha 410083, China (e-mail: hehu.mech@csu.edu.cn).}
	\thanks{Peng Shen, Jipeng Zhang, LiuJue Zhang, Linrui Xu, Zeyuan Wang, Ziyu Li, Xuezhi Cui, Wang Guo and Haifeng Li are with the School of Geosciences and Info-Physics, Central South University, Changsha 410083, China (e-mail: lihaifeng@csu.edu.cn).}
	\thanks{(Corresponding author: Hu He)}}

\markboth{IEEE Transactions on Geoscience and Remote Sensing}%
{Zhou \MakeLowercase{\textit{et al.}}: Title of Your Paper}


\IEEEoverridecommandlockouts

\IEEEaftertitletext{
	\begin{center}
		\vspace{-2.0em} 
		\includegraphics[width=0.95\textwidth]{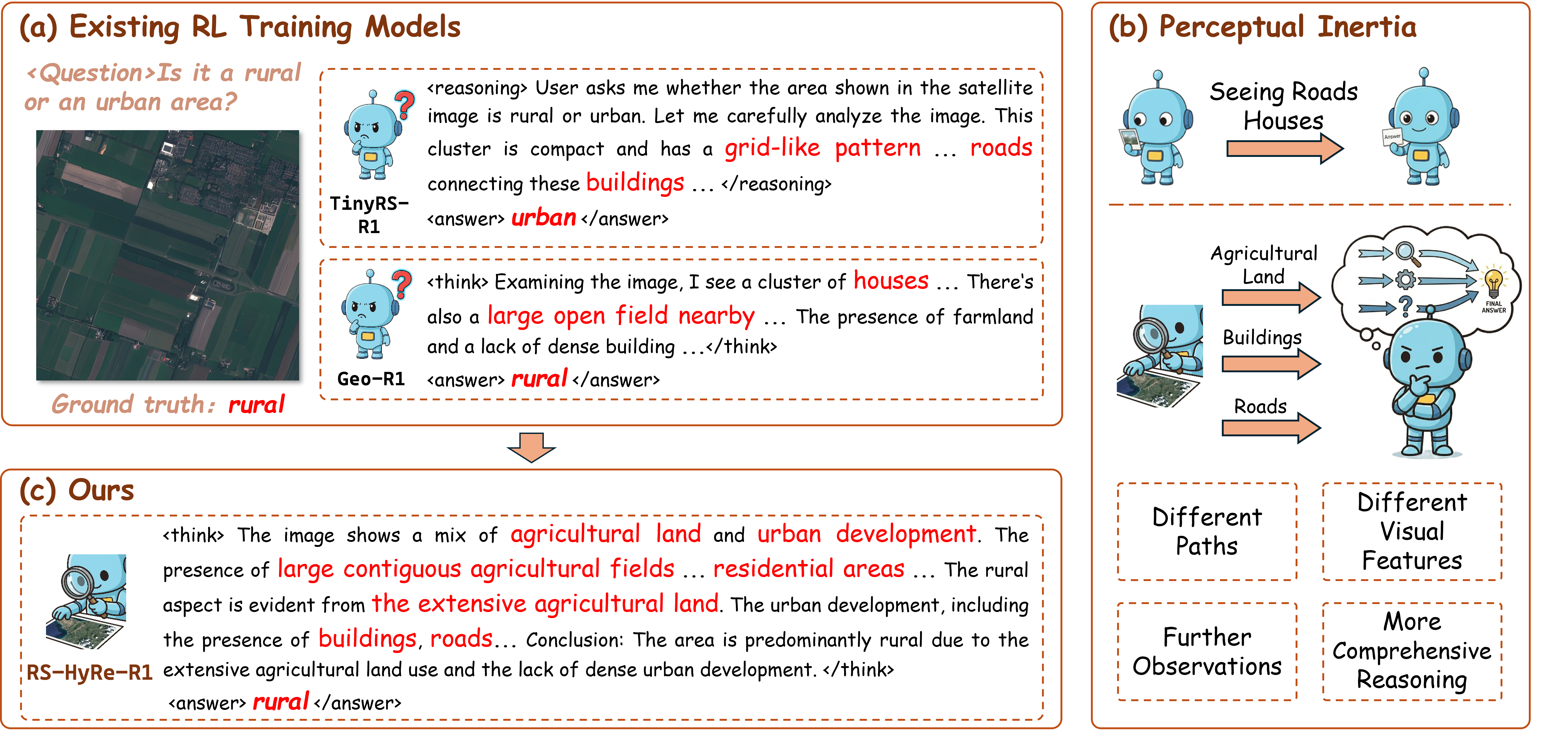}
		\par\vspace{0.5em}
		
		\refstepcounter{figure} 
		\label{fig_0}
		
		\begin{minipage}{0.95\textwidth} 
			\footnotesize
			Fig. \thefigure: Illustration of the ``perceptual inertia'' phenomenon and our proposed RS-HyRe-R1 solution.
			(a) Existing RL-driven models (e.g., TinyRS-R1, Geo-R1) often struggle with visual reasoning, exhibiting hallucinations or relying on localized salient visual cues. 
			(b) We identify this bottleneck as ``perceptual inertia'', the tendency of models to fixate on salient visual cues (e.g., buildings) while neglecting broader context. 
			(c) To overcome this, our RS-HyRe-R1 employs a hybrid reward evolutionary strategy that incentivizes ``observing more diverse''. This strategy constructs diverse and logically rigorous reasoning chains by observing more complementary visual evidence.
		\end{minipage}
		
		\vspace{0em} 
	\end{center}
}

  \maketitle

\begin{abstract}
Reinforcement learning (RL) post-training significantly enhances the performance of remote sensing vision-language models (RS-VLMs). However, when processing complex remote sensing imagery (RSI) that demands exhaustive visual scanning, models often rely on localized salient cues to rapidly deduce answers. We term this phenomenon as RL-induced ``perceptual inertia". Driven by the greedy nature of reward maximization, these models often seek rewards by quickly fitting outcomes. This greedy optimization results in dual limitations: at the cognitive level, an overreliance on specific features hinders the construction of a complete chain of evidence; at the execution level, the models struggle to flexibly shift their visual focus when handling different RS tasks. 
To overcome this optimization bias, compel the model to resume exhaustive mining of visual evidence, we propose RS-HyRe-R1, a hybrid reward mechanism for RSI understanding. Specifically, it employs a spatial reasoning activation reward to impose structural constraints for explicit visual reasoning. At the outcome-supervision level, the RS-task perception correctness reward establishes adaptive quality anchors. By dynamically calibrating answer deviations across different task paradigms, it ensures precise locking of geometric boundaries and semantic ground truths during transitions between RS tasks. Furthermore, at the reasoning process level, the visual-semantic path evolution reward is integrated to penalize repetitive reasoning patterns, thereby incentivizing the capture of neglected complementary cues to construct comprehensive evidence chains. 
Experimental results confirm that RS-HyRe-R1 effectively mitigates the ``perceptual inertia", prompting the model to generate deep reasoning sequences rich in diverse visual detail. With only 3B parameters, it achieves state-of-the-art performance across referring expression comprehension (REC), open vocabulary object detection (OVD), and visual question answering (VQA) tasks, outperforming existing models with up to 7B parameters. Notably, it demonstrates robust zero-shot generalization, surpassing the second-best model by 3.16\%, 3.97\%, and 2.72\% in VQA, OVD, and REC tasks, respectively. Code and datasets are publicly available at https://github.com/geox-lab/RS-HyRe-R1.

\end{abstract}


\section{Introduction}
\IEEEPARstart{V}{i}sion-language models (VLMs) have emerged as a crucial tool for remote sensing imagery (RSI) understanding \cite{in_2,in_3,in_4,in_5,add_in,allspark2025,peng2026,hook2024}. Recently, deep reasoning models represented by DeepSeek-R1 \cite{in_7} and Qwen3 \cite{in_8} have demonstrated the immense potential of reinforcement learning (RL) post-training paradigms \cite{in_9,in_10,rw_16}. By incorporating reward mechanisms, these models have achieved performance breakthroughs in fields such as mathematical computation and coding tasks. Inspired by this success, the RL-driven paradigm has been extended to VLMs to generate structured reasoning \cite{rw_20,new_in_1,new_in_2}.

In closely related work, existing reasoning models can be broadly categorized into two frameworks based on their reward feedback mechanisms: outcome-supervised reward models (ORMs) and process-supervised reward models (PRMs). This classification has been widely accepted in the general reasoning research community \cite{new_in_3,new_in_4}. ORMs optimize strategies using rewards derived solely from the final output (e.g., answer correctness) \cite{rw_16, rw_20,new_in_5}, whereas PRMs utilize fine-grained feedback on intermediate reasoning steps to mitigate reward sparsity \cite{new_in_4, new_in_6, new_in_7}. In the field of RS, current reasoning models primarily rely on outcome-oriented rewards \cite{com_method5,com_method6,com_method7,com_method8}. Whether employing the verifiable outcome rewards in Geo-R1 \cite{com_method5} (e.g., coordinate matching) or the ``consistency-aware rewards" introduced by GeoReason \cite{com_method8} to align reasoning with decision-making, these methods are essentially still constrained by guidance at the outcome level.

The characteristics of RSI, including broad fields of view, complex ground elements, and sparse visual cues, demand that models perform exhaustive visual scanning for accurate interpretation \cite{in_16,in_17,in_18,Cui2024AEPT,guo2026,yang2026}. However, existing outcome-guided RS reasoning models exhibit significant limitations: their inference processes often remain superficial, severely lacking in the deep exploration of fine-grained visual information. We refer to this phenomenon of visual degradation observed after RL optimization as ``perceptual inertia" (as shown in Fig. \ref{fig_0} (a) and (b)). Specifically, driven by the ``greedy" nature of RL to maximize cumulative rewards \cite{in_15,in_19,in_20}, outcome-guided models often rely on localized salient visual cues to rapidly deduce answers, rather than engaging in a detailed understanding of the RS scene. This ``perceptual inertia" results in dual limitations: at the cognitive level, an overreliance on specific salient visual features hinders the construction of a complete and rigorous chain of evidence; at the execution level, the models struggle to flexibly shift their visual focus when handling different tasks, causing feature anchoring drift and localization errors. Ultimately, these flaws lead to a degradation in generalization ability, resulting in a sharp decline in performance when faced with unseen tasks or zero-shot scenarios.

To address the aforementioned issues and mitigate ``perceptual inertia", this paper proposes RS-HyRe-R1, a hybrid reward-based RL framework for RSI understanding. We constructed a hybrid task environment encompassing referring expression comprehension (REC), open vocabulary object detection (OVD), and visual question answering (VQA), and innovatively designed a hybrid reward mechanism. First, via a spatial reasoning activation reward, we impose strict structural constraints on the reasoning process, compelling the model to generate explicit reasoning steps. Second, providing outcome-level supervision, we design an RS-task perception correctness reward specifically calibrated for REC, OVD, and VQA, ensuring precise geometric localization and semantic accuracy across tasks. Finally, at the process supervision level, we introduce a visual-semantic path evolution reward. This reward calculates the semantic similarity of different inference paths and penalizes repetitive inference patterns, thereby incentivizing the model to capture overlooked supplementary clues and construct diverse, comprehensive chains of evidence.

Experiments demonstrate that RS-HyRe-R1 achieves state-of-the-art (SOTA) performance across three core RS tasks: REC, OVD, and VQA. With superior parameter efficiency, it comprehensively surpasses existing mainstream reasoning models, including Geo-R1 and GeoReason. Qualitative analysis and response length evolution further corroborate the advantages of our framework in mitigating ``perceptual inertia". At the cognitive level, the model constructs diverse and information-rich chains of evidence rather than relying on simple, salient visual cues. At the executive level, it exhibits dynamic adaptability in visual anchoring, ensuring precise geometric localization and semantic alignment even when switching between heterogeneous tasks. These findings robustly validate the effectiveness of RS-HyRe-R1 in fostering deep visual reasoning and enhancing cross-domain generalization capabilities.

Our main contributions are summarized as follows:

\begin{itemize}
	\setlength{\itemsep}{0pt}
	\setlength{\parsep}{0pt}
	\setlength{\parskip}{0pt}
	\item [$\bullet$]
	We established a unified RL environment for diverse RS tasks, integrating three core tasks: REC, OVD, and VQA. This unification provides a standardized training platform for collaborative optimization across complex heterogeneous tasks, laying the foundation for the co-evolution of these three distinct RS tasks.
	\item [$\bullet$]
	We propose a hybrid reward-based RL framework (RS-HyRe-R1) to combat ``perceptual inertia": By simultaneously introducing hybrid reward mechanisms for both outcomes and processes, we successfully guide the model to perform exhaustive visual mining on complex RSI. While maintaining high accuracy, this framework compels the model to establish logical reasoning chains, achieving a transition from shallow feature matching to deep logical inference.
	\item [$\bullet$]
	We train RS-HyRe-R1, which achieves SOTA performance across multiple benchmarks for three RS tasks. Analyses confirm that by employing RL with a hybrid reward mechanism on RS data, the proposed method effectively mitigates ``perceptual inertia".
\end{itemize}

\section{Related Work}
\subsection{Vision-language Model for Remote Sensing}

The automated interpretation of RSI is crucial for applications such as environmental monitoring, urban planning, and disaster response \cite{rw_1,rw_2}. This necessity has propelled the rapid evolution of automated interpretation technologies. Early research primarily focused on task-specific visual grounding; for instance, addressing remote sensing visual grounding (RSVG) problems via contrastive learning or multi-scale feature fusion \cite{rw_3,rw_4}. With the emergence of Large Language Models (LLMs), researchers have endeavored to align RS visual features into the semantic space of LLMs, giving rise to RS-LVLMs such as GeoChat \cite{com_method4}, SkySense \cite{rw_6}, and RingMoGPT \cite{rw_7}. By employing supervised fine-tuning (SFT) on large-scale RS instruction datasets, these models have demonstrated robust image understanding capabilities. Subsequent works, such as SkySenseGPT \cite{rw_8} and EarthGPT \cite{rw_9}, have further expanded capabilities in multi-sensor data fusion, significantly enhancing performance across various RS benchmarks.

However, the aforementioned methods rely predominantly on SFT. While the SFT paradigm enables models to efficiently learn ``instruction-output" mapping patterns \cite{rw_10,rw_11}, it often falls short in fostering deep logical reasoning capabilities when confronting complex spatial relationships within RSI \cite{in_6,rw_12}.

\subsection{Reinforcement Learning and Reasoning Model}

To overcome the inherent limitations of SFT in handling complex logical reasoning, RL has been introduced into the post-training phase of LLMs, emerging as a key paradigm for enhancing cognitive capabilities. Seminal works such as OpenAI o1 \cite{rw_13} and DeepSeek-R1 \cite{in_7} demonstrate that learning from outcome-based feedback significantly enhances models' reasoning capabilities, particularly in mathematical derivation and code generation. Notably, the Group Relative Policy Optimization (GRPO) \cite{rw_14} algorithm introduced by DeepSeek-R1 eliminates the reliance on value networks, enabling efficient reasoning evolution.

Within this rapidly advancing domain, reasoning models are generally categorized into two frameworks based on their reward feedback mechanisms: outcome-supervised reward models (ORMs) and process-supervised reward models (PRMs) \cite{new_in_3,new_in_4}. ORMs optimize policy networks based solely on the correctness of the final output. For instance, Vision-R1 \cite{rw_16} adapts the GRPO algorithm to multi-modal tasks, employing rule-based rewards exclusively for final answer accuracy. Similarly, VLM-R1 \cite{rw_20} integrates length penalties with outcome rewards to curb reasoning redundancy while ensuring task completion. Furthermore, MM-Eureka \cite{in_9} explore self-evolved Chain-of-Thought (CoT) data, yet their RL verification remains strictly confined to the final reasoning result. In contrast, PRMs address the issue of reward sparsity by providing feedback on intermediate reasoning steps. For example, R1-VL \cite{new_in_7} introduces step-wise rewards to validate each logical deduction within the reasoning chain. Mulberry \cite{new_rw_1} utilizes collective tree search to enforce structured reasoning paths. Additionally, Visual-PRM \cite{new_rw_2} employs process reward models for step-wise verification.

However, most existing VLMs reasoning frameworks are designed for general natural images and rely on either purely outcome-based signals or generic process rewards. They often overlook the unique challenges of remote sensing imagery, such as scale variation and sparse visual cues, where coarse outcome rewards may fail to guide precise spatial reasoning.

\subsection{Remote Sensing Reasoning Model}

To adapt the complexities of RS tasks and elevate interpretation performance, researchers have increasingly integrated RL training techniques into the domain. However, most existing RS reasoning models rely on ORMs. For instance, TinyRS-R1 \cite{com_method7} pioneered compact RS-VLMs based on Qwen2VL-2B. This model employs a four-stage paradigm, encompassing pre-training, instruction tuning, CoT fine-tuning, and GRPO optimization with ORMs, to achieve exceptional performance on simple VQA tasks. Geo-R1 \cite{com_method5} advances the reasoning-first reinforcement fine-tuning (RFT) paradigm, enforcing a ``reason first, act later" protocol in few-shot scenarios to maximize the interpretability through explicit reasoning resolution. Similarly, although GeoReason \cite{com_method8} constructs a logic-driven dataset (GeoReason-Bench) and utilizes a ``SFT initialization + consistency-aware RL" strategy to enhance reasoning reliability, its feedback mechanism remains fundamentally an outcome-level constraint focused on textual alignment.

In stark contrast, this paper shifts from a pure ORMs framework to a hybrid strategy integrating both outcome-level and process-level supervision. We utilize outcome-level rewards (RS-task-aware correctness) to ensure geometric and semantic accuracy across different RS tasks, while employing process-level rewards (visual-semantic path evolution) to guide the model toward acquiring more visual information. This hybrid mechanism mitigates the “perceptual inertia” phenomenon commonly observed in RL with ORMs, compelling the model to actively engage in comprehensive visual exploration. Ultimately, we establish a hybrid reward RL framework (RS-HyRe-R1) capable of seamlessly switching between REC, OVD, and VQA tasks, demonstrating robust performance and exceptional cross-domain generalization capabilities.

\section{Task and Methodology}

\begin{figure*}[!t]
	\centering 
	\includegraphics[width=\textwidth]{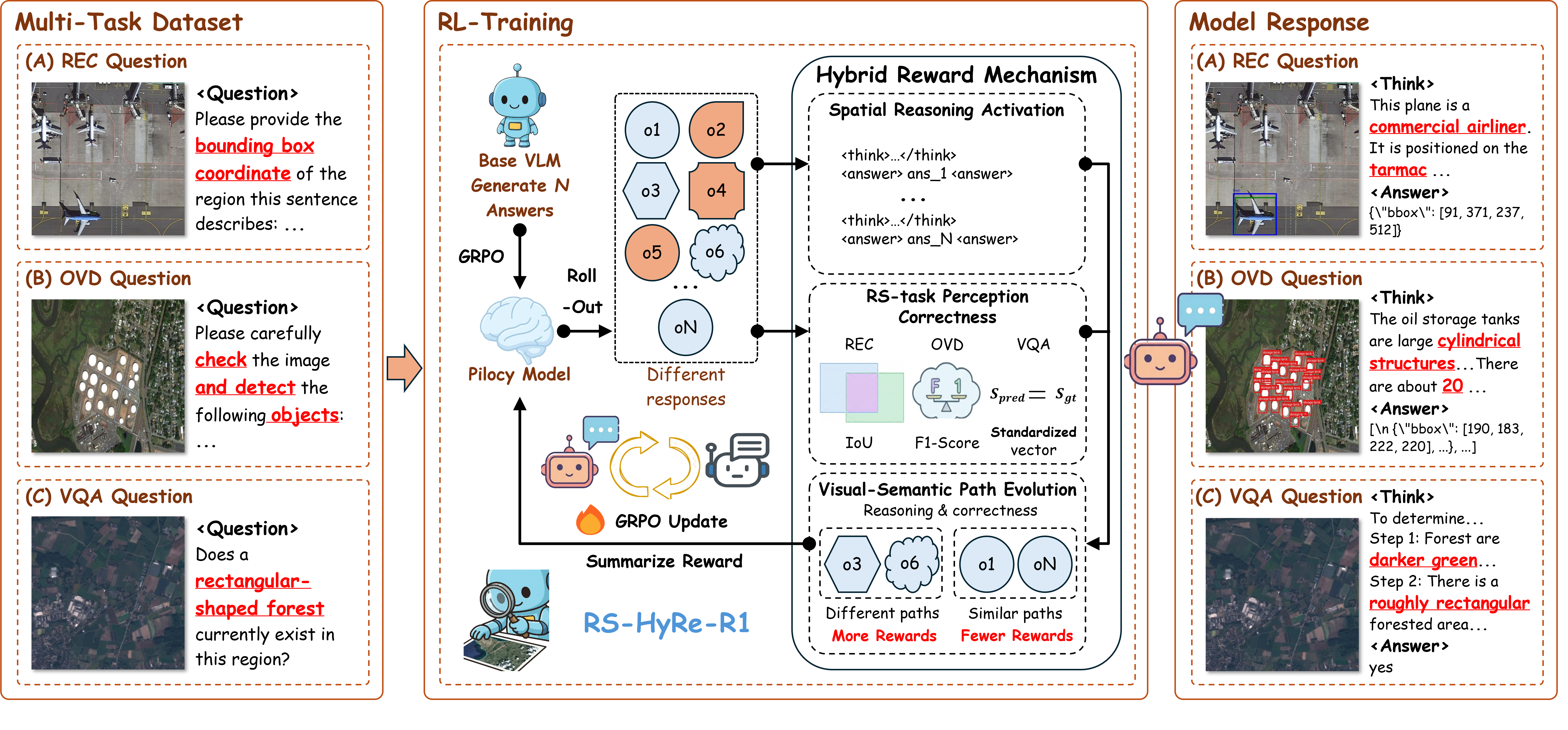}
	\caption{Overview of the proposed RS-HyRe-R1 framework.
		The pipeline begins with the construction of a RS-Task Dataset (Left), utilizing only 1,600 samples across REC, OVD, and VQA tasks. We employ GRPO for RL training (Middle), where the Policy Model generates multiple rollout answers ($o_1$ to $o_N$). The optimization is guided by the Hybrid Reward Mechanism, which evaluates outputs through three dimensions: Spatial Reasoning Activation, RS-task Perception Correctness, and Visual-Semantic Path Evolution. The rightmost section demonstrates the model's capability to generate comprehensive reasoning chains and accurate predictions across all three tasks.}
	\label{fig_1}
\end{figure*}

To mitigate the cognitive and executive limitations caused by ``perceptual inertia" in RS-VLMs during the RL post-training phase, we propose RS-HyRe-R1, a hybrid reward-based RL framework designed for RS-task that incorporates both outcome and process supervision. Our primary objective is to enable RS-VLMs to adapt to diverse tasks and incentivize it to actively explore neglected visual details. As illustrated in Fig. \ref{fig_1}, we employ GRPO as the foundational algorithm, integrated with a hybrid reward mechanism encompassing structural constraints, outcome-oriented evaluation, and process evolution. Following RL training on our curated RS-task dataset, the model demonstrates the capability to robustly perform three distinct RS tasks.

This section first elucidates the principles of the GRPO algorithm. Subsequently, we provide formal definitions for the three target RS tasks: REC, OVD, and VQA. Finally, we detail how our proposed hybrid reward mechanism adapts GRPO to address diverse task challenges in RS scenarios and mitigate ``perceptual inertia".

\subsection{Group Relative Policy Optimization (GRPO)}

We leverage the GRPO algorithm to implement RL for the post-training of vision-language models. In contrast to traditional Proximal Policy Optimization (PPO) approaches, which depend on a separate value estimator (Critic Model), GRPO eliminates this requirement. By utilizing group sampling and relative advantage estimation, GRPO significantly reduces memory overhead and improves training stability, making it particularly effective for fostering complex reasoning capabilities in large-scale models.

Specifically, for each input query $q$ (comprising an image and text prompt), the model policy $\pi_\theta$ samples a set of outputs $\{o_1,o_2,\ldots,o_G\}$, where $G$ denotes the group size. Each output $o_i$ receives a scalar reward $r_i$ based on the reward function $R\left(q,o_i\right)$. To guide optimization, GRPO computes the relative advantage $\hat{A}_i$ for each output within the group as follows:

\begin{equation}\label{equ1}
	\hat{A}_i=\frac{r_i-\mathrm{mean}\ \{r_1,\ldots,r_G\}}{\mathrm{std}\{r_1,\ldots,r_G\}}.
\end{equation}

The optimization objective of GRPO is to maximize this advantage while utilizing the KL divergence constraint to prevent the policy from deviating too far from the reference model $\pi_{ref}$. The optimization function is as follows:

\begin{equation}\label{equ2}
	\resizebox{0.9\hsize}{!}{$
	\begin{split}
		J_{\mathrm{GRPO}}\left(\theta\right) =
		& E_{\{o_i\}_{i=1}^N\sim\pi_{\theta_\text{old}}(\cdot|q)}\Biggl[
		\frac{1}{G}\sum_{i=1}^{G}
		\Biggl(
		\min\left(
		c_1\cdot\hat{A}_i,\,
		\right. \\[1ex]
		& \left.
		c_2\cdot\hat{A}_i
		\right)
		- \beta D_{\text{KL}}(\pi_\theta\parallel\pi_{\mathrm{ref}})
		\Biggr)
		\Biggr]
	\end{split}
	$},
\end{equation}
here:
\begin{equation}\label{equ3}
	c_1=\frac{\pi_\theta\left(o_i\mid q\right)}{\pi_{\theta_{\text{old}}}\left(o_i\mid q\right)}, 
\end{equation}
\begin{equation}\label{equ4}
	c_2=\mathrm{clip}\left(\frac{\pi_\theta\left(o_i\mid q\right)}{\pi_{\theta_{\text{old}}}\left(o_i\mid q\right)},1-\varepsilon,1+\varepsilon\right),
\end{equation}
where $D_{\mathrm{KL}}\left(\pi_\theta\parallel\pi_{\mathrm{ref}}\right)$ denotes the KL divergence between the current policy $\pi_\theta$ and the reference policy $\pi_{\mathrm{ref}}$, serving as a regularization term to prevent excessive deviation. The clipping mechanism within $c_2$ stabilizes training by constraining the policy update rate. In this paper, we leverage GRPO's sampling properties to not only compare performance at the outcome level but also introduce diversity competition at the inference process level, thereby evolving strategies to counteract ``perceptual inertia".

\subsection{Task Definition}

Our proposed methodology comprises three core tasks, designed to establish a unified interpretation framework for RS images understanding:

\textbf{Referring Expression Comprehension (REC)}: Given a RS image $I$ and a textual query describing a specific object (e.g.,``the small car located in the bottom-right corner adjacent to the truck"), the model is required to predict the bounding box coordinates $B_{\text{pred}}=\left[x_1,y_1,x_2,y_2\right]$ corresponding to the described object. This task demands sophisticated spatial reasoning capabilities to accurately resolve complex visual-linguistic references.

\textbf{Open Vocabulary Object Detection (OVD)}: Given a RS image $I$ and a target category list $C_{\text{list}}$, the model is tasked with identifying all instances belonging to the specified categories. The output is defined as a set of predictions $P_{\text{pred}}=\{\left(b_k,c_k\right)\}_{k=1}^N$, where $b_k$ represents the bounding box and $c_k$ denotes the corresponding category label for the $k$-th instance. This task rigorously evaluates the model's capacity for multi-object localization and semantic category alignment within complex, clutter-rich backgrounds.

\textbf{Visual Question Answering (VQA)}: Given a RS image $I$ and a natural language question $Q$, the model is required to generate a free-form textual answer $A_\text{vqa}$. This task demands the holistic integration of visual perception and semantic reasoning to interpret and respond to complex queries.

\subsection{Hybrid Reward Mechanism}
\subsubsection{Spatial Reasoning Activation Reward}

To compel the model to perform explicit visual scanning and reasoning on ground features within RSI, while ensuring the reliability of output parsing and evaluation, we introduce the spatial reasoning activation reward. This mechanism enforces explicit visual information processing by strictly verifying the presence of \text{\textless think\textgreater}...\text{\textless /think\textgreater} and \text{\textless answer\textgreater}...\text{\textless /answer\textgreater} tags in the model's output. The reward is defined as follows:

\begin{equation}\label{equ5}
	\resizebox{0.9\hsize}{!}{$
		R_{\text{srar}}(q, o) = 
		\begin{cases}
			1, & \text{if output follows the expected format} \\
			0, & \text{otherwise}
		\end{cases}
		$}.
\end{equation}

As defined in Equation (\ref{equ5}), we implement this reward as a binary function. A full reward of $1$ is assigned exclusively when the model encapsulates its reasoning process within the \text{\textless think\textgreater}...\text{\textless /think\textgreater} tag and encloses the final result within the \text{\textless answer\textgreater}...\text{\textless /answer\textgreater} tags, otherwise, the reward is $0$. This mechanism transforms the implicit CoT into an explicit constraint, compelling the model to articulate a logical reasoning path regarding the spatial relationships of ground features prior to generating final detection bounding boxes or answers.

\subsubsection{RS-task Perception Correctness Reward}

To accommodate the unique characteristics of different RS tasks, we designed adaptive quality anchors and introduced outcome-supervision for REC, OVD, and VQA to dynamically calibrate interpretation biases.

For the REC task, we adopted the Intersection over Union (IoU) between predicted and ground-truth bounding boxes as the primary metric, incorporating a segment-wise penalty mechanism to incentivize high-precision localization. The specific reward function is defined as follows:

\begin{equation}\label{equ6}
	\resizebox{0.9\hsize}{!}{$
	R_{\text{REC}}(B_{\text{pred}}, B_{\text{gt}}) = 
	\begin{cases}
		\text{IoU}(B_{\text{pred}}, B_{\text{gt}}), & \text{if } \text{IoU} \geq 0.5 \\
		0.8 \cdot \text{IoU}(B_{\text{pred}}, B_{\text{gt}}), & \text{if } 0.3 \leq \text{IoU} < 0.5 \\
		0, & \text{otherwise}
	\end{cases}
	$}.
\end{equation}
As shown in Equation (\ref{equ6}), the reward function is piecewise: if $\mathrm{IoU}\geq0.5$, the function returns the raw IoU value. For cases where $0.3\le\mathrm{IoU}<0.5$, a partial reward($0.8\times\mathrm{IoU}$) is assigned to encourage the model to progressively approximate the ground truth boundary. Conversely, if $\mathrm{IoU} < 0.3$, the reward is set to 0.

For the OVD task, we formulate it as a set prediction problem, employing a set matching metric grounded in the F1 score.
Let $P=\{p_1,\ \ldots,\ p_m\}$ denote the prediction set, where each element $p_i = (b_k, c_k)$ consists of a bounding box $b_k$ and its corresponding category label $c_k$. Similarly, let $G = \{g_1, \ldots, g_n\}$ represent the ground truth set. For any given prediction-ground truth pair $(p, g)$, we define the matching score as follows:

\begin{equation}\label{equ7}
	\resizebox{0.9\hsize}{!}{$
	M(p, g) = 
	\begin{cases}
		1.0, & \text{if } L(p) = L(g) \text{ and } \text{IoU}(p, g) \geq 0.5 \\
		0.5, & \text{if } L(p) = L(g) \text{ and } 0.3 \leq \text{IoU}(p, g) < 0.5 \\
		0, & \text{otherwise}
	\end{cases}
	$},
\end{equation}
where $L(*)$ represents the class label. Based on the maximum matching principle, we calculate the total true positive score as $TP_{\text{total}} = \sum \max M(p, g)$. The final reward is defined as:

\begin{equation}\label{equ8}
	R_{\text{OVD}} = \frac{2 \cdot P \cdot R}{P + R},
\end{equation}
where Precision $P = TP_{\text{total}} / |P|$ and Recall $R = TP_{\text{total}} / |G|$.

We not only considered IoU but also introduced a “soft matching” mechanism. This approach assigns a partial reward (0.5) even in instances of low spatial overlap ($0.3 \leq \text{IoU} < 0.5$), contingent upon the predicted category label $c_{\text{pred}}$ exhibiting semantic alignment with the ground truth.

For the VQA task, Implement a strict matching policy. The specific reward definition is as follows:
\begin{equation}\label{equ9}
	\resizebox{0.9\hsize}{!}{$
	R_{\text{VQA}}(A_{pred}, A_{gt}) = 
	\begin{cases} 
		1.0, & \text{if } \text{Norm}(A_{pred}) = \text{Norm}(A_{gt}) \\
		0, & \text{otherwise}
	\end{cases}
	$}.
\end{equation}

As defined in Equation (\ref{equ9}), a perfect score of $1$ is awarded exclusively when the predicted vector is identical to the normalized ground truth vector; all other cases yield a score of $0$.

To accommodate diverse task scenarios, we designed a dynamic task routing mechanism. This mechanism identifies specific task types by analyzing input instructions or response formats, and aggregates the evaluation feedback from sub-tasks into a unified RS-task perception correctness reward, denoted as $R_{\text{rpcr}}$. Formally, this reward is defined as follows:

\begin{equation}\label{equ9_1}
	R_{\text{rpcr}} = 
	\begin{cases} 
		R_{\text{REC}}, & \text{if } \text{task\_type} = \text{REC} \\
		R_{\text{OVD}}, & \text{if } \text{task\_type} = \text{OVD} \\
		R_{\text{VQA}}, & \text{if } \text{task\_type} = \text{VQA}
	\end{cases}.
\end{equation}

\subsubsection{Visual-Semantic Path Evolution Reward}

The visual-semantic path evolution reward serves as the key mechanism for overcoming the ``perceptual inertia". Its core logic lies in introduction ``semantic dissimilarity reward" targeting the inference process: if the model repetitively exploits a singular or stereotyped reasoning template, it receives no reward; conversely, the lower the semantic similarity between a generated path and existing paths, the higher the reward assigned. This constraint compels the policy network to escape its ``comfort zone", forcing the model to diversify its reasoning pathways. This incentivizes the model to actively explore complementary visual cues and heterogeneous logical chains.

The specific calculation process is as follows, within a single GRPO sampling iteration, we first apply terrain spatial reasoning activation reward and RS-task correctness reward. This preliminary step ensures the generation of valid reasoning paths and guarantees answer quality, preventing the model from maximizing evolutionary rewards via ungrounded, random reasoning. We restrict the evolutionary reward calculation exclusively to valid samples—specifically, those satisfying both strict format constraints ($R_{\text{srar}} > 0.99$) and high accuracy thresholds ($R_{\text{rpcr}} > 0.80$). For these qualified samples, we extract embedding vectors from the reasoning content (enclosed in \text{\textless think\textgreater} tags) to compute an intra-group cosine similarity matrix, formulated as follows:

\begin{equation}\label{equ10}
e_i=\mathrm{Embed}\left(t_i\right),  S_{ij}=\frac{e_i\cdot e_j}{|e_i||e_j|},
\end{equation}
where $S_{ij}$ denotes the semantic similarity between the $i$-th and $j$-th inference samples. To eliminate self-influence, diagonal elements are set to zero, i.e., $S_{ii}=0$. To comprehensively evaluate whether a sample has performed redundant visual inference, we calculate the average similarity between the sample and other samples within the group, defined as:

\begin{equation}\label{equ11}
{\bar{S}}_i=\frac{1}{k-1}\sum_{j}\ S_{ij}.
\end{equation}

Finally, the visual-semantic path evolution reward is defined as:
\begin{equation}\label{equ12}
R_{\text{evol},i}=1.0-\bar{S_i}.
\end{equation}
This reward mechanism establishes an inverse correlation between the similarity of reasoning responses and the assigned reward value: the lower the semantic similarity between inference paths, the higher the reward value obtained. This design explicitly incentivizes unique reasoning paths that diverge significantly from other samples. It compels the model to actively navigate diverse reasoning trajectories and exploit a broader spectrum of complementary visual cues, ultimately facilitating the construction of evidence chains that are both informationally rich and highly discriminative.

The final reward is aggregated from the three aforementioned rewards, with the overall reward defined as:
\begin{equation}\label{equ13}
R=\lambda_{\text{srar}}\cdot R_{\text{srar}}+\lambda_{\text{rpcr}}\cdot R_{\text{rpcr}}+\lambda_{\text{evol}}\cdot R_{\text{evol}}
\end{equation}
where $\lambda_{\text{srar}}$, $\lambda_{\text{rpcr}}$ and $\lambda_{\text{evol}}$ are tradeoff parameters.

\section{Experiments}
\subsection{Experimental Setup}

\textbf{Datasets}. To develop a RL environment for RS task interpretation, we constructed a RL training dataset comprising REC, OVD, and VQA. This dataset integrates data from three established open-source repositories: REC samples were extracted from VRSBench \cite{ex_ref1}, OVD data from NWPU VHR-10 \cite{ex_ref2}, and VQA samples from RSVQA-LR \cite{ex_ref3}. The resulting RS-task RL corpus consists of 1,600 samples. Table \ref{tab1} summarizes the dataset configuration and statistical details.

\begin{table}[htbp]
	\centering
	\caption{Overview of RS-task RL Training Datasets}
	\label{tab1}
	\begin{tabular}{ccccc}

		\toprule[1pt] 
		Task & Source Dataset & Samples & Images & Definition \\
		\midrule
		REC & VRSBench & 260 & 254 & image-query-box \\
		OVD & NWPU VHR-10 & 650 & 150 & image-query-boxes\\
		VQA & RSVQA-LR & 690 & 274 & image-query-answer \\
		\midrule
		\multicolumn{2}{c}{RS-Task} & 1600 & 678  & \\

		\bottomrule[1pt] 
	\end{tabular}
\end{table}

\textbf{Model and Training Details}. We employ Qwen2.5-VL-3B-Instruct \cite{basemodel} as the base model to demonstrate its performance on lightweight architectures. Our implementation utilizes the Easy-R1\footnote{https://github.com/hiyouga/EasyR1} codebase \cite{easyr1}. Unless otherwise specified, we retain default hyperparameters without manual tuning. We incorporate thought prompting into the RL framework, utilizing GRPO as the post-training paradigm. GRPO is configured with 15 generations per rollout. The hybrid reward mechanism weights configured as $\lambda_\text{srar}=0.1, \lambda_\text{rpcr}=0.7, \text{ and } \lambda_{evol}=0.2$. These parameter values were selected based on experimental experience. Training was conducted on four A800 GPUs for approximately 72 to 96 hours (40 epochs), yielding a unified model with a single set of weights for all three tasks, eliminating the need for task-specific fine-tuning.

\textbf{Benchmark}. To comprehensively evaluate the model's capabilities, we employ a multi-dimensional validation strategy. We assess RS task performance across REC, OVD, and VQA using three distinct RS benchmarks: RSVQA-LR-test \cite{ex_ref3}, NWPU VHR-10 \cite{ex_ref2}, and VRSBench-test \cite{ex_ref1}. Table \ref{tab2} details the statistical characteristics of these test sets.

\begin{table}[htbp]
	\centering
	\caption{Overview of Test Datasets}
	\label{tab2}
	\begin{tabular}{cccc}
		\toprule[1pt] 
		Task & Source Dataset & Samples & Images \\
		\midrule
		REC & VRSBbench-test & 16159  & 9318 \\
		OVD & NWPU VHR-10 & 2424 & 500 \\
		VQA & RSVQA-LR-test & 10003 & 1001\\
		\bottomrule[1pt]
	\end{tabular}
\end{table}

To evaluate the model's zero-shot generalization capability, we extended our assessment to unseen benchmarks, specifically DIOR-RSVG \cite{in_6} for the REC task, RSOD \cite{ex_ref4} for the OVD task, and VRSBench-VQA \cite{ex_ref1} for the VQA task. Table \ref{tab3} summarizes the statistics for these test sets.
\begin{table}[htbp]
	\centering
	\caption{Overview of Zero-shot Test Datasets}
	\label{tab3}
	\begin{tabular}{cccc}
		\toprule[1pt]
		Task & Source Dataset & Samples & Images\\
		\midrule
		REC & DIOR-RSVG & 7422  & 3372 \\
		OVD & RSOD & 936 & 936 \\
		VQA & VRSBench-VQA & 37409 & 9349 \\
		\bottomrule[1pt] 
	\end{tabular}
\end{table}

\textbf{Metrics}. We evaluate experimental performance using standard, widely accepted metrics tailored to each task: Intersection over Union (IoU) for REC, Mean Average Precision (mAP) for OVD, and Pass@1 accuracy for VQA.

\textbf{Baseline}. To rigorously evaluate the efficacy of our proposed RS-HyRe-R1, we compare it against three categories of SOTA methods. First, we utilize the official Zero-shot Baselines of Qwen2.5-VL \cite{basemodel} checkpoints (3B and 7B) to assess the foundational performance of general-purpose VLMs without domain adaptation. Second, in the SFT category, we compare against our own Qwen2.5-VL-SFT \cite{basemodel} (fine-tuned on our 1,600-samples dataset) and GeoChat \cite{com_method4}, a leading RS-VLM fine-tuned on a massive 36k+ instruction dataset, alongside other notable SFT baselines such as LLaVA-1.5 \cite{com_method1}, Mini-Gemini \cite{com_method2}, and MiniGPT-v2 \cite{com_method3}. Finally, we benchmark against recent RL-driven Models that integrate RL for reasoning, including TinyRS-R1 \cite{com_method7}, Geo-R1 \cite{com_method5}, GeoReason \cite{com_method8} and R1-VL \cite{new_in_7}(PRMs), and task-specific variants like VLM-R1-OVD/REC \cite{rw_20} and Geo-R1-OVD/REC \cite{com_method6}.

\subsection{Performance Analysis}
\subsubsection{REC Task}
\textbf{Task Evaluation}. We evaluate REC performance using Acc@$t$, defined as the percentage of samples where the IoU between the predicted and ground truth bounding boxes exceeds the threshold $t$. In this study, we report the stringent metrics of Acc@0.5 and Acc@0.7. The highest-performing results are highlighted in bold.

\textbf{Results Analysis}. As shown in Table \ref{tab4}, RS-HyRe-R1 achieves SOTA performance on the VRSBench-test dataset. It significantly surpasses the SFT+RL trained TinyRS-R1 by 27.42\% on Acc@0.5 and more than triples its performance on the stricter Acc@0.7 metric (32.25\% vs. 10.07\%). Notably, our framework outperforms both the generic pure PRMs baseline R1-VL and the pure ORMs baseline VLM-R1-REC. Despite utilizing a smaller 3B backbone, RS-HyRe-R1 outperforms 7B-parameter models like GeoReason by 9.69\% on Acc@0.5 and 8.75\% on Acc@0.7. It even edges out the task-specific variant Geo-R1-REC, demonstrating that our framework dramatically enhances precise target localization without compromising task-specific accuracy. Furthermore, RS-HyRe-R1 outperforms fully fine-tuned models such as GeoChat, establishing a new benchmark for data efficiency. This validates that our hybrid reward mechanism, integrating outcome-level correctness with process-level path evolution, effectively mitigates ``perceptual inertia''. By compelling the model to engage in deep visual scanning, the strategy significantly enhances the precise localization of target geometric boundaries while reducing reliance on massive annotated data.

\begin{table}[htbp]
	\centering
	\caption{Quantitative Performance Comparison of REC Task on the VRSBench-test Dataset}
	\label{tab4}
	\footnotesize
	\resizebox{\linewidth}{!}{
		\begin{tabular}{lccc}
			\toprule[1pt]
			\multirow{2}{*}{Method} & \multirow{2}{*}{Base LLM} & \multicolumn{2}{c}{Overall} \\
			\cmidrule(lr){3-4}
			& & Acc@0.5 & Acc@0.7 \\
			\midrule
			\multicolumn{4}{c}{Zero-shot Baseline} \\
			\midrule
			Qwen2.5-VL 3B \cite{basemodel} & Qwen2.5-3B & 16.89 & 7.36 \\
			Qwen2.5-VL 7B \cite{basemodel} & Qwen2.5-7B & 41.28 & 23.32 \\
			\midrule
			\multicolumn{4}{c}{RS-task Dataset Fine-tune (1600 samples)} \\
			\midrule
			Qwen2.5-VL-SFT & Qwen2.5-3B & 35.39 & 18.58 \\
			\midrule
			\multicolumn{4}{c}{Full Amount Fine-tune (36,313 samples)} \\
			\midrule
			LLaVA-1.5 \cite{com_method1} & Vicuna1.5-7B & 41.60 & 13.60 \\
			Mini-Gemini \cite{com_method2} & Gemma-7B & 30.10 & 6.80 \\
			MiniGPT-v2 \cite{com_method3} & Vicuna1.5-7B & 35.80 & 16.80 \\
			GeoChat \cite{com_method4} & Vicuna1.5-7B & 49.80 & 19.90 \\
			\midrule
			\multicolumn{4}{c}{RL Train} \\
			\midrule
			TinyRS-R1 \cite{com_method7} & Qwen2-2B & 23.94 & 10.07 \\
			Geo-R1 \cite{com_method5} & Qwen2.5-7B & 42.03 & 17.62 \\
			GeoReason \cite{com_method8} & Qwen2.5-7B & 41.67 & 23.50 \\
			R1-VL \cite{new_in_7} & Qwen2-7B & 35.02 & 19.28 \\
			VLM-R1-OVD \cite{rw_20} & Qwen2.5-3B & 33.51 & 17.68 \\
			VLM-R1-REC \cite{rw_20} & Qwen2.5-3B & 38.77 & 21.18 \\
			Geo-R1-OVD \cite{com_method6} & Qwen2.5-3B & 38.22 & 20.90 \\
			Geo-R1-REC \cite{com_method6} & Qwen2.5-3B & 49.60 & 29.77 \\
			RS-HyRe-R1 & Qwen2.5-3B & \textbf{51.36} & \textbf{32.25} \\
			\bottomrule[1pt]
		\end{tabular}
	}
\end{table}

\subsubsection{OVD Task}

\textbf{Task Evaluation}. For the OVD task, we employed the standard COCO-style mAP metric on the NWPU VHR-10-val dataset, which explicitly excludes the 150 image samples designated for training in the original NWPU VHR-10 dataset. We focus on two primary metrics: mAP@0.5 and mAP@[0.5:0.95]. The mAP@0.5 metric measures baseline detection accuracy at an IoU threshold of 0.5, reflecting fundamental localization capability. In contrast, mAP@[0.5:0.95] averages precision across IoU thresholds ranging from 0.5 to 0.95 (in 0.05 increments). By imposing stricter requirements on bounding box overlap, this metric provides a more rigorous and comprehensive assessment of the model's localization precision across varying levels of stringency.

\textbf{Results Analysis}. The quantitative assessment in Table \ref{tab5} highlights RS-HyRe-R1's superior OVD capabilities. It achieves a mAP@[0.5:0.95] of 0.2600 and a mAP@0.5 of 0.5225. The experimental results highlight significant advantages over both similarly sized and larger-scale RL-driven models. Specifically. It more than doubles the performance of GeoReason and R1-VL, indicating that the inference trajectories generated by our proposed hybrid reward mechanism are more effective than the logical consistency constraints of GeoReason and the general process reward mechanism of R1-VL. Even when compared to GEO-R1-OVD, a variant optimized solely for detection, RS-HyRe-R1 achieves a performance leap of approximately 18\% on the mAP@0.5. This substantial gain strongly validates that the proposed framework effectively achieves more precise target acquisition and localization in OVD scenarios by dynamically calibrating cross-task deviations and incentivizing diverse visual reasoning within the model.

\begin{table}[htbp]
	\centering
	\caption{Quantitative Performance Comparison of OVD Task on the NWPU VHR-10-val Dataset}
	\label{tab5}
	\footnotesize
	\resizebox{\linewidth}{!}{
		\begin{tabular}{lccc}
			\toprule[1pt]
			\multirow{2}{*}{Method} & \multirow{2}{*}{Base LLM} & \multicolumn{2}{c}{Overall} \\
			\cmidrule(lr){3-4}
			& & mAP@[0.5:0.95] & mAP@0.5 \\
			\midrule
			\multicolumn{4}{c}{Zero-shot Baseline} \\
			\midrule 
			Qwen2.5-VL 3B \cite{basemodel} & Qwen2.5-3B & 0.0386 & 0.0948 \\
			Qwen2.5-VL 7B \cite{basemodel} & Qwen2.5-7B & 0.1044 & 0.1539 \\
			\midrule
			\multicolumn{4}{c}{RS-task Dataset Fine-tune (1600 samples)} \\
			\midrule
			Qwen2.5-VL-SFT & Qwen2.5-3B & 0.1592 & 0.3474 \\
			GeoChat \cite{com_method4} & Vicuna1.5-7B & 0.1731 & 0.3509 \\
			\midrule
			\multicolumn{4}{c}{RL Train} \\
			\midrule
			TinyRS-R1 \cite{com_method7} & Qwen2-2B & 0.0873 & 0.1605 \\
			Geo-R1 \cite{com_method5} & Qwen2.5-7B & 0.1561 & 0.3358 \\
			GeoReason \cite{com_method8} & Qwen2.5-7B & 0.1205 & 0.2916 \\
			R1-VL \cite{new_in_7} & Qwen2-7B & 0.1012 & 0.2590 \\
			VLM-R1-OVD \cite{rw_20} & Qwen2.5-3B & 0.0740 & 0.1901 \\
			VLM-R1-REC \cite{rw_20} & Qwen2.5-3B & 0.0603 & 0.1658 \\
			Geo-R1-OVD \cite{com_method6} & Qwen2.5-3B & 0.1887 & 0.3428 \\
			Geo-R1-REC \cite{com_method6} & Qwen2.5-3B & 0.1067 & 0.2612 \\
			RS-HyRe-R1 & Qwen2.5-3B & \textbf{0.2600} & \textbf{0.5225} \\
			\bottomrule[1pt]
		\end{tabular}
	}
\end{table}

\subsubsection{VQA Task}

\textbf{Task Evaluation}. For the VQA task, we adopt Top-1 Accuracy (Pass@1) as the primary evaluation metric on the RSVQA-LR-test dataset. This metric measures the proportion of samples where the model's highest-ranked prediction yields an exact match with the ground truth. It serves as a rigorous indicator of the model's logical reasoning capabilities and precision when interpreting complex visual-linguistic scenarios.

\textbf{Results Analysis}. The quantitative analysis in Table \ref{tab6} reveals the breakthrough performance of RS-HyRe-R1 on the VQA task. RS-HyRe-R1 achieves a Pass@1 accuracy of 59.87\%. It substantially outperforms both the generic pure PRMs baseline R1-VL (48.21\%) and domain-adapted pure ORMs frameworks such as GeoReason (50.32\%), GEO-R1 (46.88\%) and VLM-R1-OVD (49.16\%).  This significant margin, nearly 10\% over 7B-parameter reasoning models, proves that relying solely on generic step-wise verification or outcome-level textual consistency is inadequate for resolving complex RS visual queries. It proves that the RS-HyRe-R1 can capture rich visual details to support complex logical reasoning, outperforming larger-scale models despite its compact size.

\begin{table}[htbp]
	\centering
	\caption{Quantitative Performance Comparison of VQA Task on the RSVQA-LR-test Dataset}
	\label{tab6}
	\footnotesize 
	\renewcommand{\arraystretch}{1.2} 
	\setlength{\tabcolsep}{8pt} 
	
	\begin{tabular}{lcc}
		\toprule[1pt]
		Method & Base LLM & Pass@1 \\
		\midrule
		\multicolumn{3}{c}{Zero-shot Baseline} \\
		\midrule 
		Qwen2.5-VL 3B \cite{basemodel} & Qwen2.5-3B & 39.27 \\
		Qwen2.5-VL 7B \cite{basemodel} & Qwen2.5-7B & 46.21 \\
		\midrule
		\multicolumn{3}{c}{RS-task Dataset Fine-tune (1600 samples)} \\
		\midrule
		Qwen2.5-VL-SFT & Qwen2.5-3B & 44.63 \\
		GeoChat \cite{com_method4} & Vicuna1.5-7B & 56.62  \\
		\midrule
		\multicolumn{3}{c}{RL Train} \\
		\midrule
		TinyRS-R1 \cite{com_method7} & Qwen2-2B & 51.73 \\
		Geo-R1 \cite{com_method5} & Qwen2.5-7B & 46.88  \\
		GeoReason \cite{com_method8} & Qwen2.5-7B & 50.32 \\
		R1-VL \cite{new_in_7} & Qwen2-7B & 48.21 \\
		VLM-R1-OVD \cite{rw_20} & Qwen2.5-3B & 49.16 \\
		VLM-R1-REC \cite{rw_20} & Qwen2.5-3B & 44.86 \\
		Geo-R1-OVD \cite{com_method6} & Qwen2.5-3B & 46.29  \\
		Geo-R1-REC \cite{com_method6} & Qwen2.5-3B & 44.66  \\
		RS-HyRe-R1 & Qwen2.5-3B & \textbf{59.87}  \\
		\bottomrule[1pt]
	\end{tabular}
\end{table}

Experimental results across these three distinct RS tasks demonstrate that RS-HyRe-R1 achieves outstanding SOTA performance, fully validating its robust RS-task generalization capabilities. Despite utilizing a compact 3B-parameter architecture and fine-tuning on only 1,600 RS-task samples, RS-HyRe-R1 comprehensively outperforms larger-scale baseline models, generic PRMs, and domain-adapted pure ORMs across the REC, OVD, and VQA tasks. This cross-scale performance superiority powerfully underscores the value of the proposed hybrid reward framework. By integrating outcome-level correctness (ORMs) with process-level path evolution (PRMs), the hybrid reward mechanism effectively mitigates both the ``perceptual inertia'' induced by greedy optimization and the ``feature drift'' commonly observed during heterogeneous RS-task transitions. Consequently, the framework endows the model with pixel-level localization precision for detection and referential tasks, while ensuring logical reasoning completeness for the VQA task. Ultimately, this establishes dual robustness at both the cognitive and executive levels within complex RS environments.

\subsection{Evaluation of Zero-Shot Generalization across Datasets}

To thoroughly evaluate the model's generalization ability under non-homogeneous data distributions, we conducted cross-dataset zero-shot testing for the REC, OVD, and VQA tasks. In this experimental setup, all comparative models underwent limited supervised fine-tuning on our constructed 1,600-samples RS-task dataset and were subsequently deployed for inference on completely unseen target domains. We used the DIOR-RSVG, RSOD-test, and VRSBench-VQA datasets to validate the models' generalization capabilities on the REC, OVD, and VQA tasks, respectively.

\textbf{Results Analysis}. The experimental results in Table \ref{tab7} further validate the robust zero-shot generalization capability of RS-HyRe-R1 under data distribution shifts. Despite undergoing training on a limited number of source domain samples, our model maintains significant performance advantages when directly transferred to completely unseen target domains. Specifically, in the REC transfer task (VRSBench-REC $\to$ DIOR-RSVG), RS-HyRe-R1 achieves a Acc@0.5 of 46.85\% and a Acc@0.7 of 31.84\%. These scores comprehensively surpass both the larger-parameter GeoChat model (44.13\% and 30.62\%) and the Geo-R1-REC baseline. Furthermore, in the OVD zero-shot evaluation (NWPU VHR-10 $\to$ RSOD-test), RS-HyRe-R1 demonstrates superior robustness in unseen target detection, achieving an mAP@[0.5:0.95] of 0.2350. This significantly outperforms the fully fine-tuned GeoChat model (0.1953) and the base model (0.1778). Similarly, in VQA cross-domain testing (RSVQA $\to$ VRSBench-VQA), it achieves a Pass@1 accuracy of 45.13\%, exceeding GeoChat by approximately 3.2\% and the base fine-tuned Qwen2.5-VL-SFT by nearly 15\%. These results strongly suggest that the proposed hybrid reward mechanism enables the model to construct more generalized visual-language mapping logic and a comprehensive visual scanning strategy, prompting the model to focus on the geometric and semantic essence of target objects rather than specific dataset-induced biases. Consequently, the model acquires robust transfer learning capabilities, facilitating accurate reasoning on non-homogeneous data.

\begin{table}[htbp]
	\centering
	\caption{Quantitative Performance Comparison of Cross-Dataset Zero-Shot Generalization for REC, OVD, and VQA Tasks}
	\label{tab7}
	\footnotesize 
	\renewcommand{\arraystretch}{1.2} 
	\setlength{\tabcolsep}{2pt}
	
	\begin{tabular}{l c c c c}
		\toprule
		\multicolumn{1}{c}{\multirow{4.5}{*}{\textbf{Method}}} & 
		\multicolumn{2}{c}{\begin{tabular}{@{}c@{}}
				\textbf{VRSBench-REC} \\ 
				$\downarrow$ \\ 
				\textbf{DIOR-RSVG}
		\end{tabular}} & 
		\begin{tabular}{@{}c@{}}
			\textbf{NWPU VHR-10} \\ 
			$\downarrow$ \\ 
			\textbf{RSOD-test}
		\end{tabular} &
		\begin{tabular}{@{}c@{}}
			\textbf{RSVQA} \\ 
			$\downarrow$ \\ 
			\textbf{VRSBench-VQA}
		\end{tabular} \\
		
		\cmidrule(lr){2-3} \cmidrule(lr){4-4} \cmidrule(lr){5-5}
		
		& Acc@0.5 & Acc@0.7 & mAP@[0.5:0.95] & Pass@1 \\
		\midrule
		Qwen2.5-VL-SFT & 28.22 & 13.76 & 0.1778 & 30.15 \\
		GeoChat & 44.13 & 30.62 & 0.1953 & 41.97 \\
		Geo-R1-REC & 43.20 & 29.32 & 0.1004 & 40.22\\
		RS-HyRe-R1 & \textbf{46.85} & \textbf{31.84} & \textbf{0.2350} & \textbf{45.13} \\
		\bottomrule
	\end{tabular}
\end{table}

\begin{figure*}[p]
	\centering
	\setlength{\tabcolsep}{3pt}
	
	\begin{tabular}{m{0.02\textwidth} m{0.96\textwidth}}
		\centering\textbf{(a)} & 
		\includegraphics[width=\linewidth, height=\textheight, keepaspectratio]{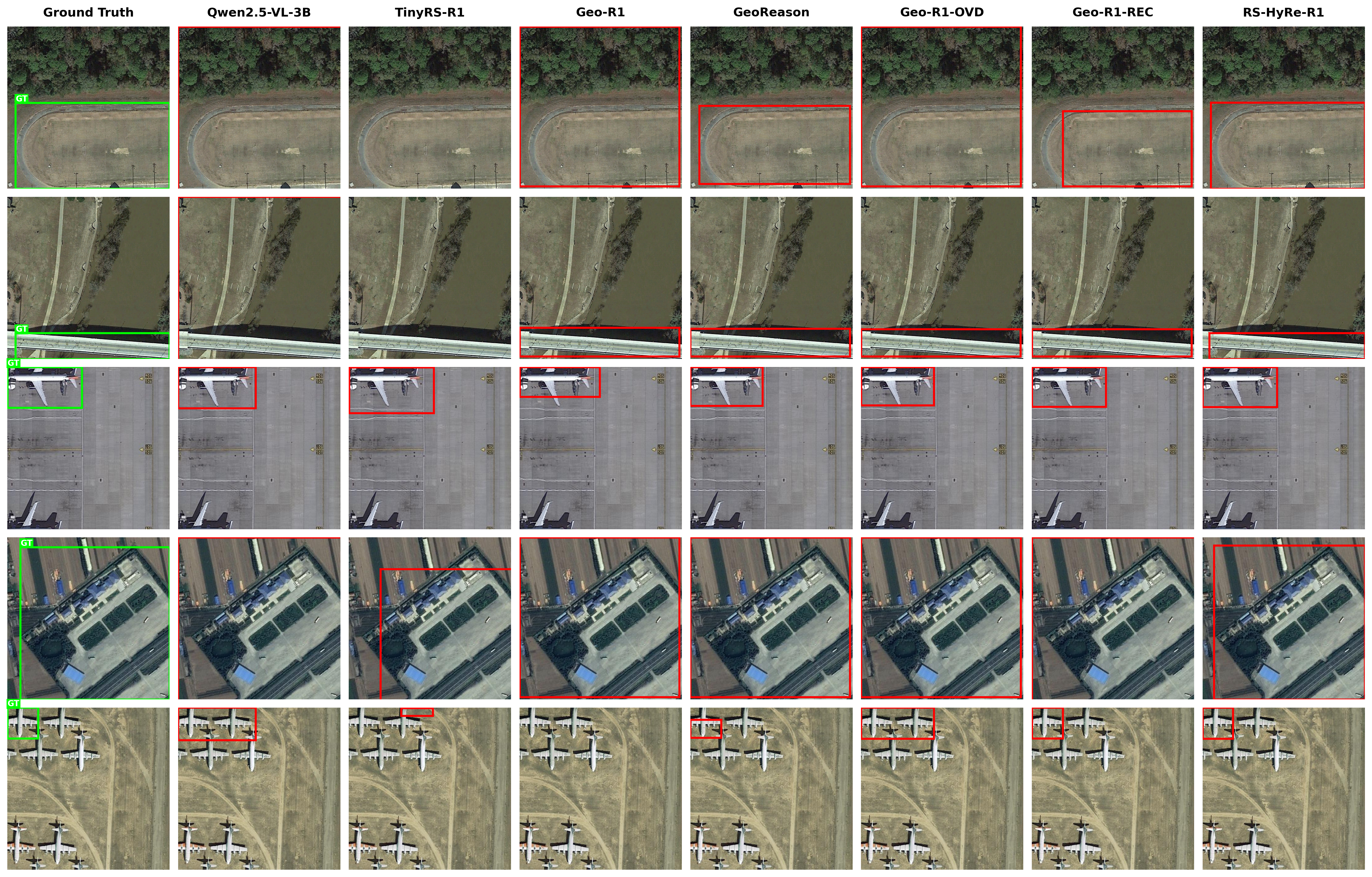} \\ 
		
		& \\[-1ex] 
		
		\centering\textbf{(b)} & 
		\includegraphics[width=\linewidth, height=\textheight, keepaspectratio]{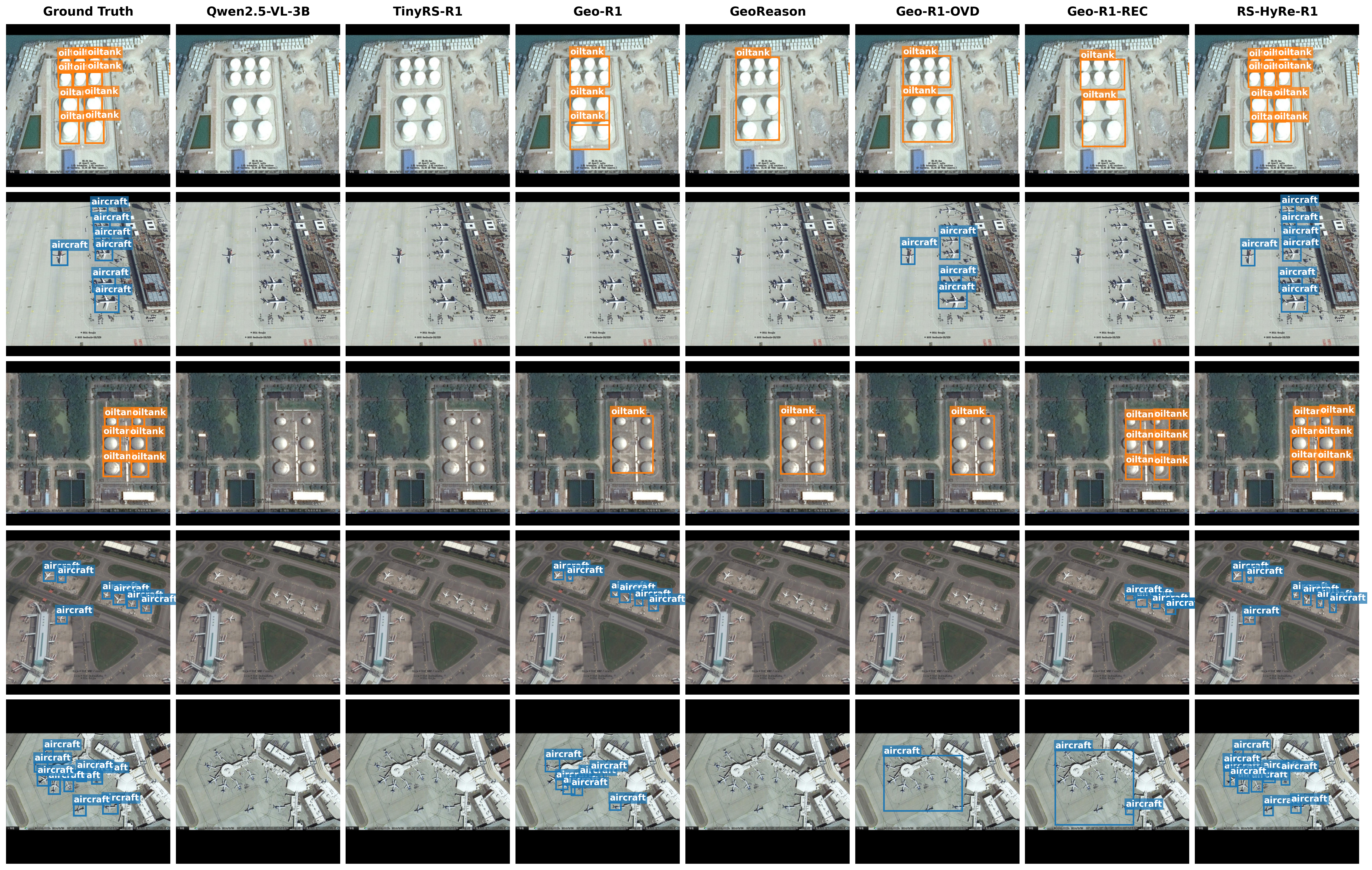} 
	\end{tabular}
	
	\caption{Qualitative comparison results with six models (including five RL training models). (a) Visualization of REC Task Results on the VRSBench-test Dataset. (b) Visualization of OVD Task Results on the RSOD-test Dataset. The proposed RS-HyRe-R1 consistently outperforms other baselines.}
	\label{fig:visual_comparison}
\end{figure*}  

\subsection{Qualitative Analysis}

We conducted a comprehensive qualitative evaluation of the model's performance on the REC and OVD tasks, as illustrated in Fig. \ref{fig:visual_comparison}. The visualization results provide concrete evidence of RS-HyRe-R1's superiority over five competing RL-driven models.

In the REC task (Fig. \ref{fig:visual_comparison}(a)), RS-HyRe-R1 exhibits exceptional localization precision. Compared to baselines like TinyRS-R1 and GeoReason, which often generate loose bounding boxes or drift away from the target, our model consistently produces tight, accurate predictions that align closely with the Ground Truth. Notably, RS-HyRe-R1 achieves localization accuracy comparable to, and in complex scenes even slightly surpassing, the specialized Geo-R1-REC model, verifying that our unified training does not compromise task-specific precision.

In the OVD task (Fig. \ref{fig:visual_comparison}(b)), which challenges the model's ability to handle dense, multi-scale objects, RS-HyRe-R1 demonstrates robust detection capabilities. As observed in the dense storage tank and aircraft scenarios, existing models often suffer from high omission rates (false negatives) for small-scale targets. In contrast, RS-HyRe-R1 successfully identifies a significantly higher number of small objects, outperforming SOTA baselines including Geo-R1-OVD. This qualitative superiority confirms that our evolutionary reward mechanism effectively compels the model to perform meticulous visual scanning, thereby overcoming ``perceptual inertia" and ensuring comprehensive target discovery in cluttered RS environments.

\subsection{Analysis of Reasoning Expansion}

\begin{figure}[H]
	\centering
	\includegraphics[width=\linewidth]{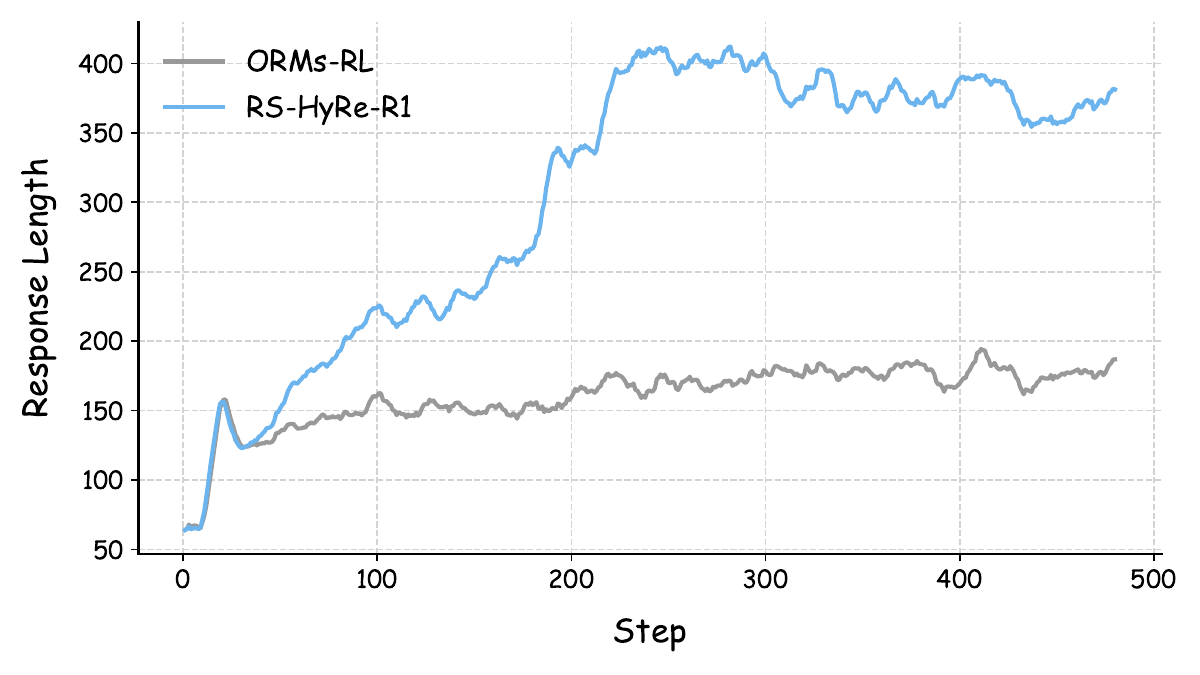}
	\caption{Comparison of response length trajectories. Unlike the ORMs-RL baseline (gray), which converges to 150--180 tokens, RS-HyRe-R1 (blue) surges to 350--400 tokens. This significant increase indicates that the proposed hybrid reward evolutionary strategy performs more in-depth visual reasoning.}
	\label{fig_2}
\end{figure}

To investigate cognitive behavioral shifts during RL, Fig. \ref{fig_2} compares the average response length evolution of RS-HyRe-R1 against an ORMs-RL baseline (guided solely by correctness reward), both trained on the identical 1,600 RS-task samples. Initially (first 20 steps), both models generate short responses of approximately 60 tokens. However, their trajectories soon diverge. The ORMs-RL model's length sluggishly plateaus between 150 and 180 tokens, suggesting that outcome-only feedback encourages shallow semantic analysis and reliance on prominent visual features. In sharp contrast, RS-HyRe-R1's response length surges after step 50, ultimately stabilizing in the high range of 350 to 400 tokens by step 250. This substantial disparity, more than double that of the baseline, does not represent redundant information but rather a significantly extended CoT. Driven by the visual-semantic path reward, our model is compelled to overcome ``perceptual inertia'' and actively leverage deeper visual details for multi-hop reasoning. This ``short-to-long'' evolution strongly validates that our strategy successfully transitions the model from simple outcome prediction to deep visual-logical reasoning.

\subsection{RL Training Stability Analysis}

\begin{figure}[H]
	\centering
	\includegraphics[width=\linewidth]{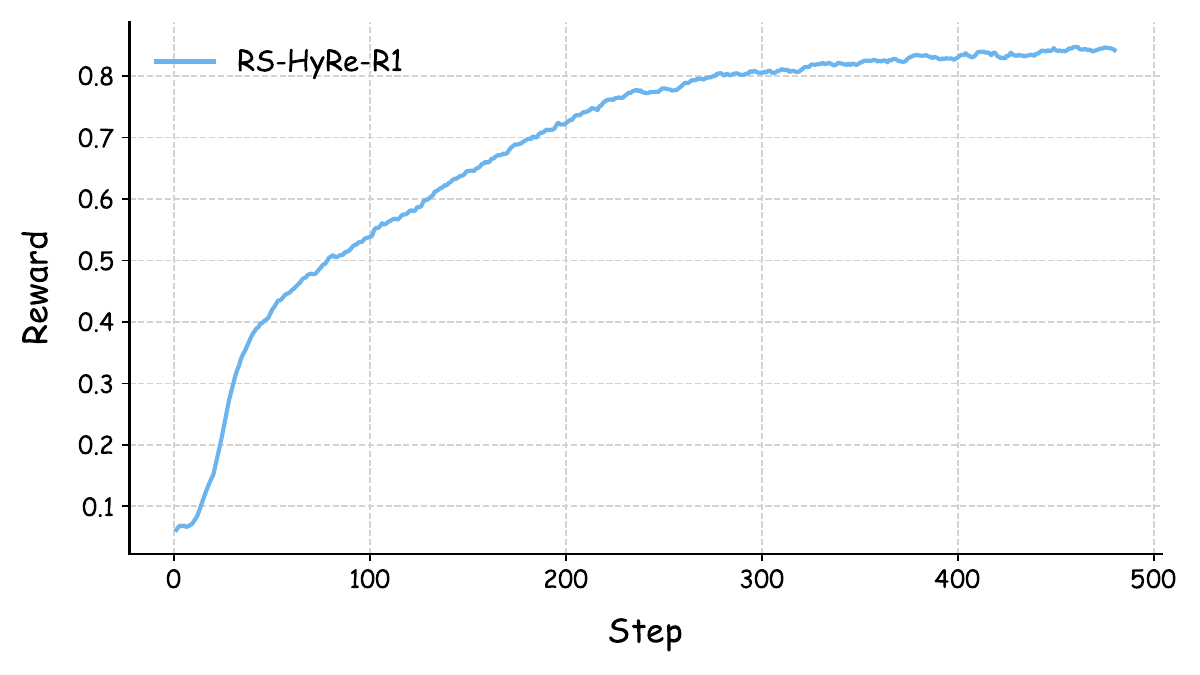}
	\caption{The overall reward curve for the RL training. The curve illustrates the steady ascent of the total reward for RS-HyRe-R1 throughout the training process. This stable upward trajectory demonstrates the effectiveness of our hybrid reward mechanism in providing a robust learning signal, effectively guiding the model toward optimal performance without exhibiting training instability or collapse.}
	\label{fig_3}
\end{figure}

Fig. \ref{fig_3} illustrates the overall reward curve of the RS-HyRe-R1 model during the RL training process. We observe a smooth and consistent upward trajectory, where the average reward steadily increases from an initial value of approximately 0.06 and converges around 0.85 after step 400. This stable rise strongly indicates the effectiveness of our reconstructed reward strategy designed for RS data. The convergence behavior suggests that our proposed multi-source hybrid reward mechanism successfully mitigates reward hacking and provides a stable, robust learning signal. This enables the model to learn efficiently across heterogeneous RS hybrid datasets without training collapse.

\subsection{Ablation Study}

\begin{figure*}[t]
	\centering
	\begin{minipage}{0.32\linewidth}
		\centering
		\includegraphics[width=\linewidth]{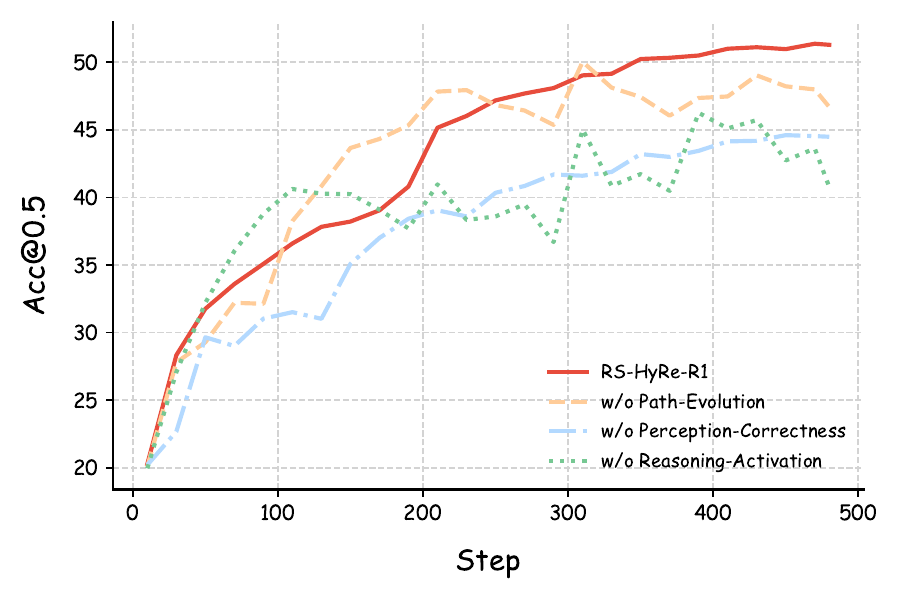}
		\caption*{(a) REC Task}
		\label{fig:ab_rec}
	\end{minipage}
	\hfill 
	\begin{minipage}{0.32\linewidth}
		\centering
		\includegraphics[width=\linewidth]{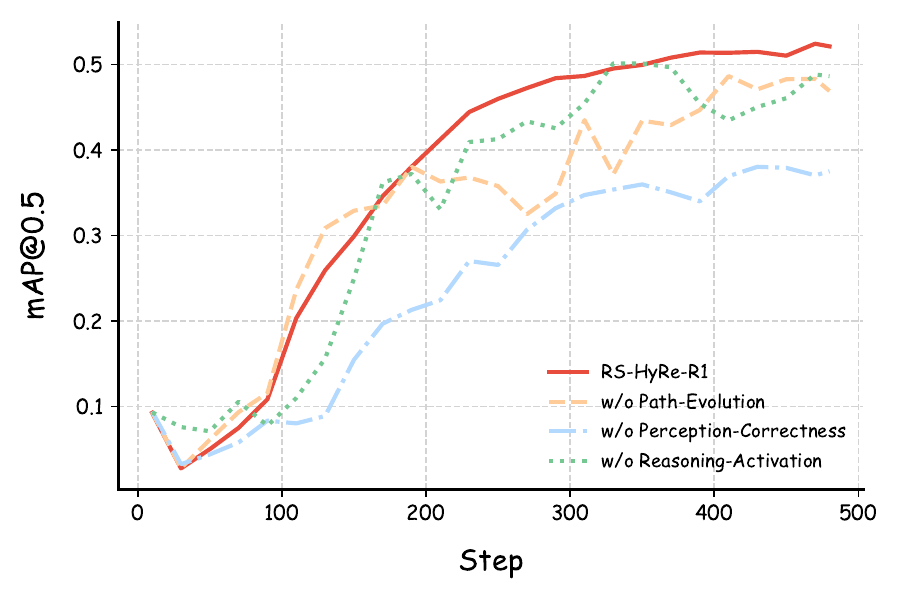}
		\caption*{(b) OVD Task}
		\label{fig:ab_ovd}
	\end{minipage}
	\hfill  
	\begin{minipage}{0.32\linewidth}
		\centering
		\includegraphics[width=\linewidth]{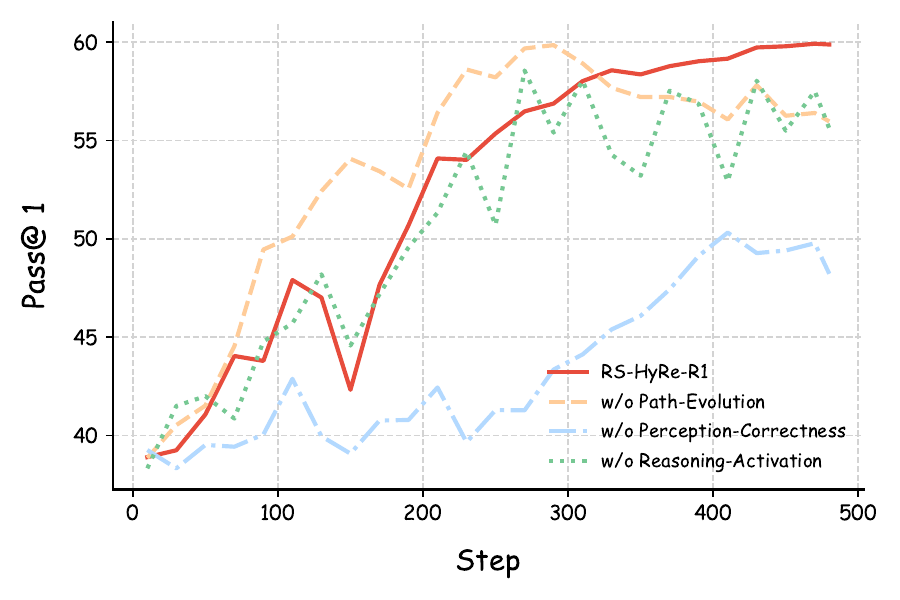}
		\caption*{(c) VQA Task}
		\label{fig:ab_vqa}
	\end{minipage}
	
	\caption{Ablation study for different rewards.
		The comparison curves of (a) Acc@0.5 for REC, (b) mAP@0.5 for OVD, and (c) Pass@1 for VQA during the training process. 
		The complete model (red solid line) achieves optimal performance and robustness; removing the perceptual correctness reward (blue dashed-dotted line) results in a significant drop in performance; removing the reasoning activation reward (green dashed line) causes significant training fluctuations; and while removing the path evolution reward (orange dashed line) initially relies on salient features to adapt quickly to the reward, its subsequent performance improvement is hindered by ``perceptual inertia".}
	\label{fig_4}
\end{figure*}

The proposed hybrid reward mechanism, integrating both outcome-level and process-level supervision, successfully mitigates ``perceptual inertia". To validate this assertion, as illustrated in Fig. \ref{fig_4}, we conducted an ablation study by removing each reward component individually. Specifically, the model lacking the ``visual-semantic path evolution reward" (w/o Path-Evolution) shows a higher performance gain during the early stages of training. However, as training progresses, its later performance plateaus and is eventually overtaken by the full model. This confirms that without this reward constraint, the model tends to rapidly fit rewards using localized salient visual cues for quick convergence; yet, due to the lack of deep exploration into diverse visual information, it ultimately falls into local optima.

Furthermore, experimental results reveal the critical roles of the other components. The model lacking the ``RS-task perception correctness reward" (w/o Perception-Correctness) performs the worst across all metrics, strongly proving that outcome-level supervision is indispensable as the core of the reward mechanism for ensuring geometric and semantic alignment. Meanwhile, the model without the ``spatial reasoning activation reward" (w/o Reasoning-Activation) exhibits severe performance fluctuations during training, demonstrating that this reward ensures the stability of reasoning through structural constraints, preventing randomness in logical trajectories. In summary, the full model achieves the best performance across all tasks, fully demonstrating that the three reward mechanisms are indispensable in synergistically promoting deep visual reasoning.

\section{Conclusion}

This paper addresses the critical challenge of ``perceptual inertia" in VLMs during the RL post-training phase for RS interpretation. Specifically, existing outcome-guided reasoning models tend to rely on localized salient visual cues for rapid deduction, neglecting the exhaustive visual scanning required for complex RS imagery. To tackle this, we propose RS-HyRe-R1, a hybrid reward-based RL framework tailored for comprehensive RS-task adaptation.

To counteract this optimization bias and encourage comprehensive visual grounding, we constructed a unified training environment integrating REC, OVD, and VQA tasks, driven by a novel hybrid reward mechanism. Specifically, we enforce explicit structural reasoning via the spatial reasoning activation reward. At the outcome level, we guarantee localization precision and semantic accuracy across task transitions through the RS-task perception correctness reward. At the process supervision level, we introduce the visual-semantic path evolution reward to incentivize the model to actively discover complementary visual cues to construct diverse, evidence-based reasoning chains.

Extensive experimental results strongly validate the effectiveness of our model. RS-HyRe-R1 successfully overcomes the ``perceptual inertia" inherent in traditional RL, significantly increasing the utilization of visual information during deep logical inference. With only 3B parameters, the model comprehensively outperforms larger-scale fine-tuned models (e.g., GeoChat), as well as existing outcome-guided reasoning models (e.g., Geo-R1 and GeoReason) and process-guided reasoning models (such as R1-VL) across the REC, OVD, and VQA tasks. Crucially, the framework endows the model with dual robustness at both cognitive and executive levels, enabling dynamic visual anchoring and precise geometric localization even when facing distinct RS tasks and zero-shot datasets.

Future work will explore the application of this evolutionary framework to multi-modal data such as multi-spectral and SAR imagery, and investigate its scalability with larger datasets, aiming to build a more generalizable \cite{zhao2026,he2026,wang2025}, and interpretable intelligent RS interpretation system.

\bibliography{RS-HyRe-R1}

\begin{IEEEbiography}[{\includegraphics[width=1in,height=1.25in,clip,keepaspectratio]{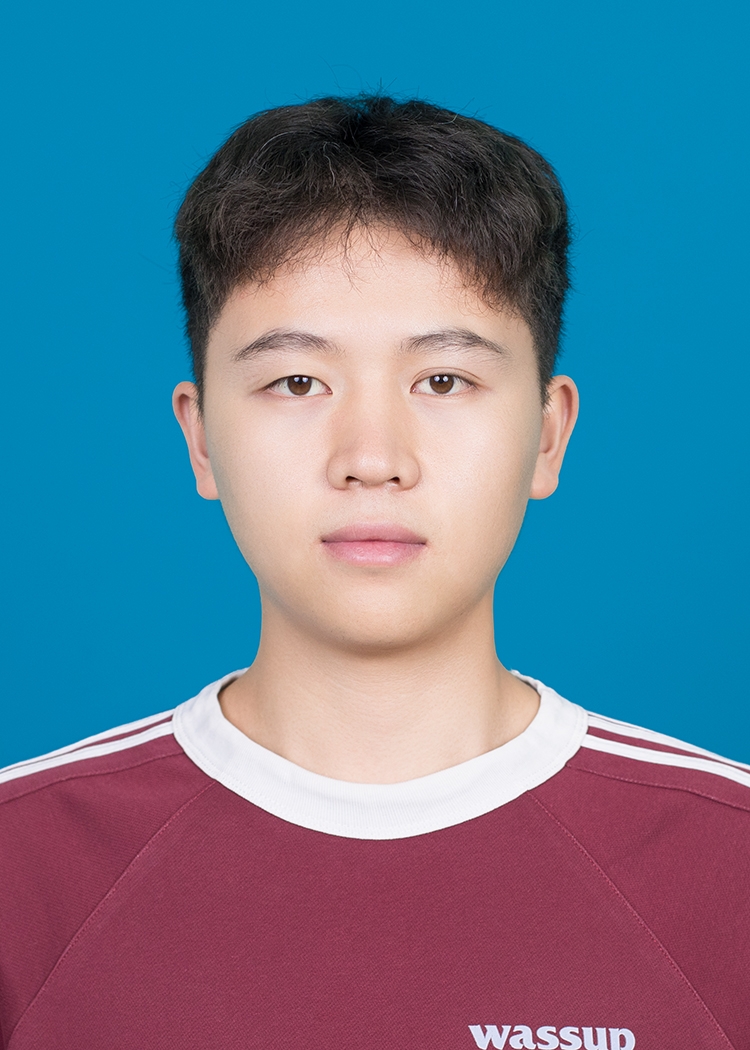}}]{Gaozhi Zhou}
	received the Master's degree in Electronic Information from Kunming University of Science and Technology, Kunming, China, in 2025. He is currently pursuing the Ph.D. with the School of Mechanical and Electrical Engineering, Central South University, Changsha, China. His research interests include multimodal reasoning models, computer vision, and remote sensing image understanding.
\end{IEEEbiography}

\begin{IEEEbiography}[{\includegraphics[width=1in,height=1.25in,clip,keepaspectratio]{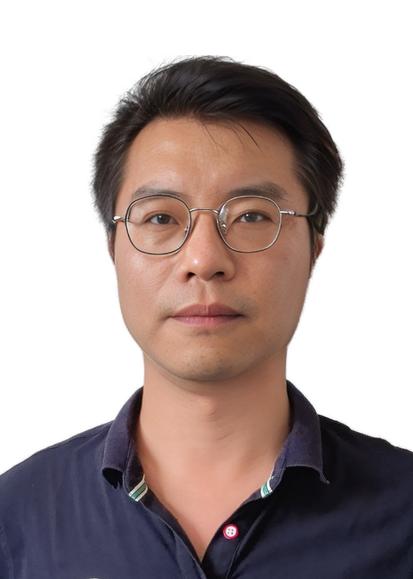}}]{Hu He}
	received a B.Eng degree in Microelectronic Manufacturing Engineering from Central South University, Changsha, China, in 2008, and a Ph.D. degree in computer science and electrical engineering from the Queensland University of Technology, Brisbane, QLD, Australia, in 2014. He is an Associate Professor with Central South University, Changsha, China. He has authored more than 50 articles in international journals and conferences. His current research interests include electronic packaging and reliability, AI for engineering, and precision measurement technology.
\end{IEEEbiography}

\begin{IEEEbiography}[{\includegraphics[width=1in,height=1.25in,clip,keepaspectratio]{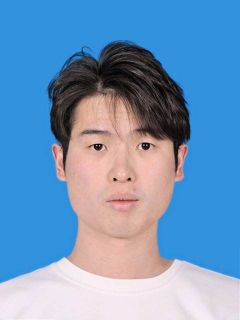}}]{Peng Shen}
	received the M.D. degree from University of Science and Technology Beijing, China, in 2024. He is currently pursuing the Ph.D. degree in the School of Geosciences and Info-Physics, Central South University. His research interests include time series and spatio-temporal data mining.
\end{IEEEbiography}

\begin{IEEEbiography}[{\includegraphics[width=1in,height=1.25in,clip,keepaspectratio]{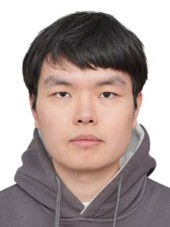}}]{Jipeng Zhang}
	is a Ph.D. student in the Department of Geomatics and remote sensing at Central South Universitty. His research interests include embodied agents in UAV and memory systems of agents.
\end{IEEEbiography}

\begin{IEEEbiography}[{\includegraphics[width=1in,height=1.25in,clip,keepaspectratio]{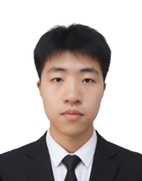}}]{Liujue Zhang}
	received the Master's degree in Mechanical and Automation Engineering from The Chinese University of Hong Kong, Hong Kong, China, in 2025. He is currently pursuing the Ph.D. with the School of Geo-Sciences and Info-Physics, Central South University, Changsha, China. His research interests include multimodal reasoning models, computer vision, and remote sensing image understanding.
\end{IEEEbiography}

\begin{IEEEbiography}[{\includegraphics[width=1in,height=1.25in,clip,keepaspectratio]{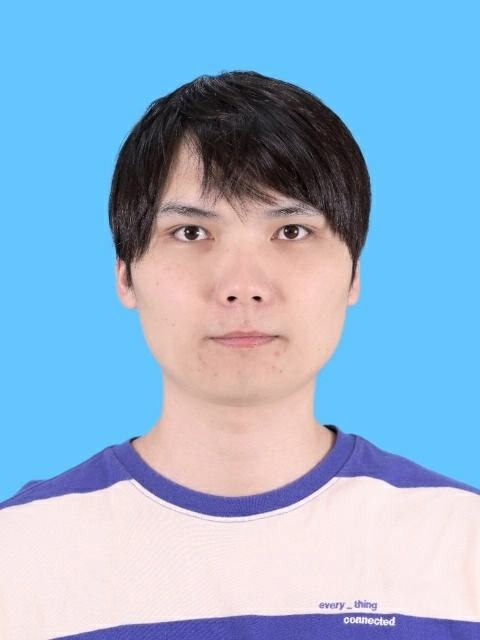}}]{Linrui Xu}
	is currently pursuing the Ph.D. degree with Central South University, Changsha, China. His research interests include computer vision, multimodal reasoning large language models, and remote sensing image understanding.
\end{IEEEbiography}

\begin{IEEEbiography}[{\includegraphics[width=1in,height=1.25in,clip,keepaspectratio]{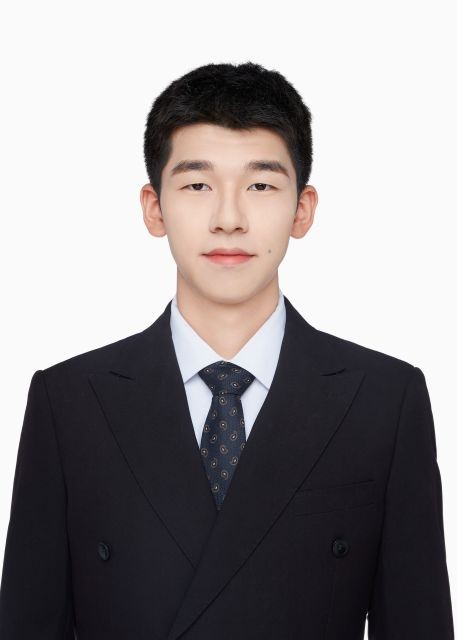}}]{Zeyuan Wang}
	received the Bachelor’s degree in Safety Engineering from the School of Resources and Safety Engineering, Central South University, Changsha, China, in 2025. He is currently pursuing the Master’s degree in Remote Sensing Science and Technology with the School of Geosciences and Info-Physics, Central South University, Changsha, China. His research interests include remote sensing agent tool learning, multimodal remote sensing agent memory.
\end{IEEEbiography}

\begin{IEEEbiography}[{\includegraphics[width=1in,height=1.25in,clip,keepaspectratio]{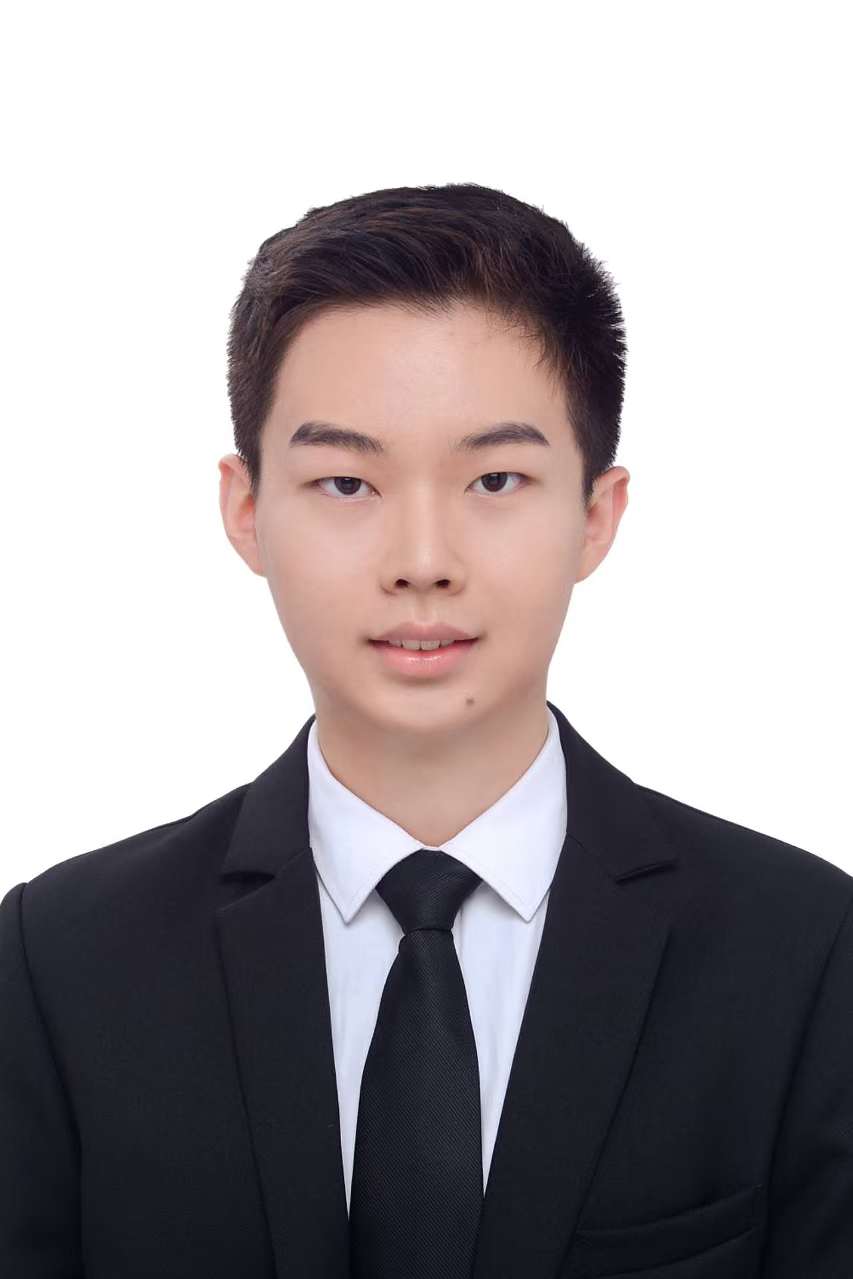}}]{Ziyu Li}
	received the B.S. degree in Geographic Information Science from Central South University, Changsha, Hunan, China, in June 2025. He is currently pursuing the Ph.D. degree in Surveying and Mapping Science and Technology at Central South University. His research interests include reasoning techniques for remote sensing large language models, multimodal geospatial understanding, and intelligent analysis of Earth observation data.
\end{IEEEbiography}

\begin{IEEEbiography}[{\includegraphics[width=1in,height=1.25in,clip,keepaspectratio]{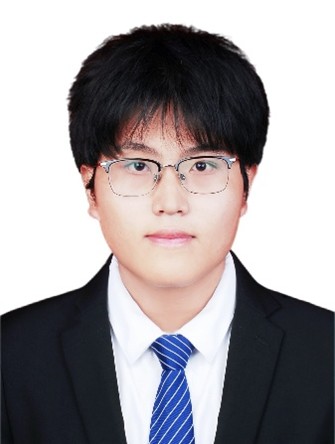}}]{Xuezhi Cui}
	received the B.S. degree in Geographic Information Science from Central South University, Changsha, Hunan, China, in June 2025. He is currently pursuing a Ph.D. degree in Surveying and Mapping Science and Technology at Central South University. His research interests include intelligent interpretation of remote sensing imagery, vision-language models, and continual fine-tuning of pre-trained models.
\end{IEEEbiography}

\begin{IEEEbiography}[{\includegraphics[width=1in,height=1.25in,clip,keepaspectratio]{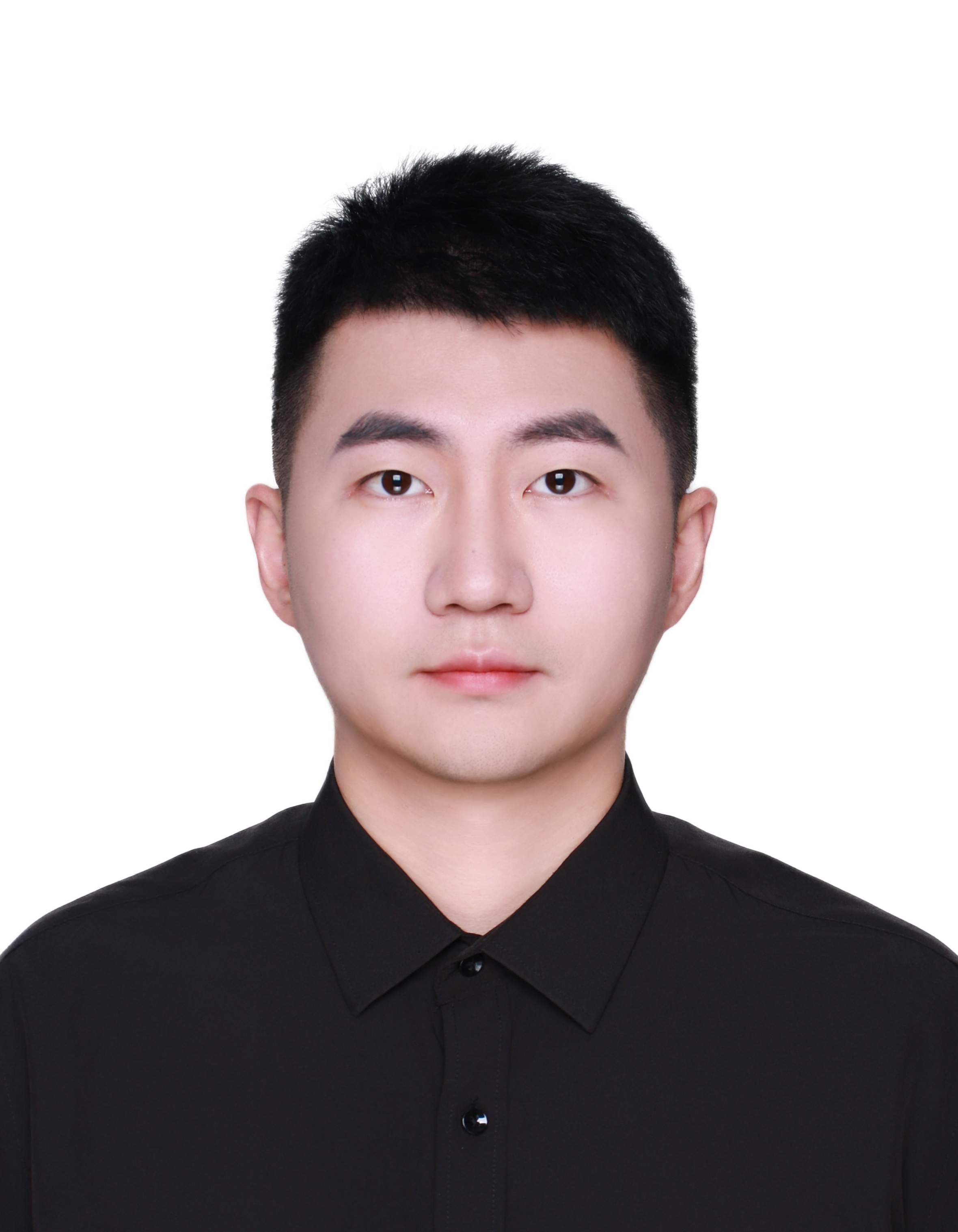}}]{Wang Guo}
	received the M.Sc. degree in the Interactive Media from University College Cork, Cork, Ireland, in 2019 and is currently pursuing the Ph.D. degree in Central South University, Changsha, China. His research interests include memory model, lifelong learning and remote sensing image processing.
\end{IEEEbiography}

\begin{IEEEbiography}[{\includegraphics[width=1in,height=1.25in,clip,keepaspectratio]{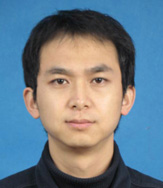}}]{Haifeng Li}
	received a master's degree in transportation engineering from the South China University of Technology, Guangzhou, China, in 2005, and a Ph.D. degree in photogrammetry and remote sensing from Wuhan University, Wuhan, China, in 2009. He is currently a professor at the School of Geosciences and Info-Physics, Central South University, Changsha, China. He was a research associate with the Department of Land Surveying and Geo-Informatics, The Hong Kong Polytechnic University, Hong Kong, in 2011, and a visiting scholar with the University of Illinois at Urbana-Champaign, Urbana, IL, USA, from 2013 to 2014. He has authored over 30 journal papers. His current research interests include geo/remote sensing big data, machine/deep learning, and artificial/brain-inspired intelligence. He is a reviewer for many journals.
\end{IEEEbiography}

\end{document}